\documentclass[journal]{IEEEtran}
\usepackage[utf8]{inputenc}
\usepackage{times,enumerate} 
\usepackage[usenames,dvipsnames,svgnames,table]{xcolor}
\usepackage{subcaption}
\usepackage{graphicx}
\usepackage[noadjust]{cite}

\usepackage{rotating}
\usepackage{csquotes}
\usepackage{amsmath}
\usepackage{bm}
\usepackage{tabularx}
\usepackage{stfloats}
\usepackage{threeparttable}
\usepackage{url}
\usepackage{soul}
\usepackage{threeparttable}
\soulregister\cite7
\soulregister\ref7
\soulregister\pageref7
\usepackage{amssymb}
\usepackage{soul}
\usepackage{footnote}
\usepackage{colortbl}
\usepackage{soul}
\usepackage{multirow}
\usepackage{pifont}
\usepackage{algorithmic}
\usepackage{booktabs,xcolor,colortbl}
\usepackage{color}
\usepackage{alltt}
\usepackage{hyperref}
\usepackage{enumerate}
\usepackage{siunitx}
\usepackage{epstopdf}
\usepackage{bbding}
\usepackage{pbox}
 \usepackage{amsmath}  
 \usepackage{amssymb}   
 \usepackage{amsthm}
 \usepackage[dvipsnames]{xcolor}
 
\usepackage{algorithm}

\newcommand{\probP}{\text{I\kern-0.15em P}}
\newcommand{\probE}{\text{I\kern-0.15em E}}

\begin{document}

\title{Give and Take: Federated Transfer Learning for Industrial IoT Network Intrusion Detection}

\author{
\IEEEauthorblockN{Lochana Telugu Rajesh\IEEEauthorrefmark{1}, Tapadhir Das\IEEEauthorrefmark{2}
Raj Mani Shukla\IEEEauthorrefmark{1}, and 
Shamik Sengupta\IEEEauthorrefmark{3}\\}
\IEEEauthorblockA{\IEEEauthorrefmark{1}School of Computing and Information Science, Anglia Ruskin University, UK\\}
{\IEEEauthorrefmark{2}Department of Computer Science, 
University of the Pacific, USA\\}
{\IEEEauthorrefmark{2}Department of Computer Science and Engineering, 
University of Nevada, Reno, USA\\
Email: lt689@student.aru.ac.uk, raj.shukla@aru.ac.uk, tdas@pacific.edu, ssengupta@unr.edu}
}
\IEEEoverridecommandlockouts
\IEEEpubid{\makebox[\columnwidth]{978-1-5386-5541-2/18/\$31.00~\copyright2018 IEEE \hfill}
\hspace{\columnsep}\makebox[\columnwidth]{ }}

\maketitle

\begin{abstract}
The rapid growth in Internet of Things (IoT) technology has become an integral part of today's industries forming the Industrial IoT (IIoT) initiative, where industries are leveraging IoT to improve communication and connectivity via emerging solutions like data analytics and cloud computing. Unfortunately, the rapid use of IoT has made it an attractive target for cybercriminals. Therefore, protecting these systems is of utmost importance. In this paper, we propose a federated transfer learning (FTL) approach to perform IIoT network intrusion detection. As part of the research, we also propose a combinational neural network as the centerpiece for performing FTL. The proposed technique splits IoT data between the client and server devices to generate corresponding models, and the weights of the client models are combined to update the server model. Results showcase high performance for the FTL setup between iterations on both the IIoT clients and the server. Additionally, the proposed FTL setup achieves better overall performance than contemporary machine learning algorithms at performing network intrusion detection.
\end{abstract}


\section{Introduction}
\begin{figure}[t]
\centerline{\includegraphics[width=\columnwidth]{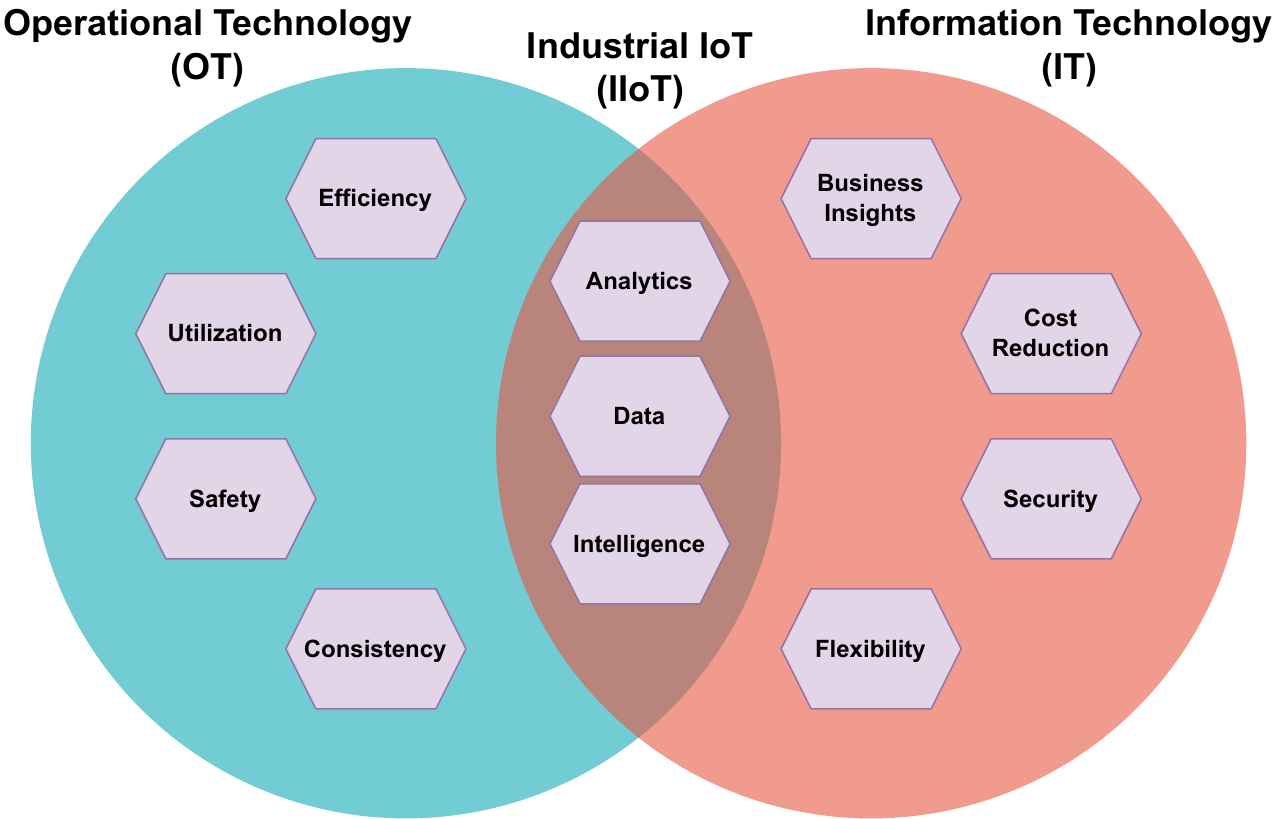}}
\caption{Industrial Internet of Things}
\label{Fig:iiot}
\end{figure}

The Internet of Things (IoT) has become an integral part of our lives as it connects people, things, and processes flawlessly. At a very low cost and with minimal human intervention, the physical world can meet the digital world. Certain instances can include smart watches that use sensors to keep track of your heartbeat rate and footstep count, motion sensor lights that toggle on when you enter or exit a room, or even smart transit systems that can provide passengers with predicted wait times for public transportation. IoT can be an energy-saving resource that promotes automation to help the public. Cisco stated that by the end of 2023, there will be more than three devices connected to the internet for every human on the planet \cite{cisco2020cisco}. Industrial IoT (IIoT) is the intersection of information technology (IT) and operational technology (OT). IT utilizes computing devices to process electronic data while OT uses programmable devices that gather environmental information through repeated interactions. IIoT forms an integral part of today's industrial applications as the physical infrastructures of various industries like agriculture, healthcare, and power, can be monitored using this technology. An illustration of IIoT is provided in Figure \ref{Fig:iiot}. 

A major challenge to the adoption of IIoT is data security. IIoT consists of a network of interconnected devices. Hence, if a device in the network is compromised, all other devices in the network could be put in jeopardy.  Therefore, it is essential to secure the sensitive data of all IIoT devices through measures such as secured data ports, transmission channels, and the network. Kaspersky reported that within the first six months of 2021, 1.5 billion IoT devices had been attacked, an increase of 639 million from the previous year \cite{cyrus2021iot}. A major hindrance to protecting these networks is the progressively increasing number of users, resulting in the generation of sizable and substantial data. Most of this generated data is private and sensitive, requiring it to be protected from cybercriminals. Unlike traditional IT, where there are considerable cybersecurity tools to protect the data, IIoT data is not protected from cyber attacks by simply using traditional cybersecurity methods. 

With the rise of computational capabilities and faster processing power, the usage of machine learning (ML) for performing IoT network intrusion detection can be a valuable option for protecting IIoT data from being attacked. However, certain limitations are currently present with existing ML-based network intrusion detection systems (NIDS). The first limitation lies in the scarcity of IIoT data. ML algorithms require high performance to be effective as a network security mechanism, which requires tremendous amounts of highly-dimensional data. Unfortunately, many laws such as the General Data Protection Regulation (GDPR) do not allow the collection of data without the consent of data subjects. Collection of IoT device data can be challenging as this data can contain sensitive and private user data. For instance, a personal voice assistant might contain overheard personal information from a user's conversation. Additionally, another limitation is the willingness to share the data with relevant individuals and organizations. To maintain confidentiality, organizations prohibit the sharing of their data with the outside world. In some organizations, data sharing is restricted to a subset of individuals who work directly with the data and is kept hidden from other employees. To ensure security, certain practices restrict data sharing across international borders. Therefore, finding a data source collected from IIoT devices is a non-trivial task due to the restrictions placed on data-sharing practices by organizations. 

The second limitation lies with the traditional ML algorithms that have been investigated to create ML-based NIDS for IIoT. Due to the high-dimensional nature of IoT data, existing ML algorithms tend to lose their efficiency. Also, most IoT data is collected from heterogenous sources like water sensors and soil moisture sensors. This can make it difficult to create a comprehensive ML algorithm capable of analyzing the different robust statistics and variances of the data sources \cite{al2021review}. Additionally, developing ML or deep learning (DL) techniques to perform IIoT intrusion detection is challenging as new attack types and vectors are getting discovered frequently. These new attack types are not being effectively detected by existing ML and DL methods and can also require more computation power to operate \cite{sharma2021optimal}.  

To address the issue of data availability, federated learning (FL) has been seen as an effective mechanism. FL helps to build personalized models for IIoT devices that do not violate user and organizational privacy through novel training methods. In addition, FL provides a privacy protection scheme that utilizes the computing resources of the IIoT device to train the models, thereby preventing information leakage during transmissions \cite{zhang2021survey}. Similarly, to address the limitation of heterogeneous IIoT data sources, transfer learning (TL) can be an effective solution. TL focuses on transferring knowledge on previously learned IIoT data to another domain to reduce computational complexity and redundant training of ML models \cite{zhuang2020comprehensive}. Through the usage of FL and TL, ML-NIDS can be generated that protect IoT user privacy and also can accommodate ML algorithms that can account for heterogeneous data sources. 

In this paper, we propose a novel federated transfer learning (FTL) approach to perform IIoT network intrusion detection. As the centerpiece for our proposed approach, we propose a novel combinational neural network (Combo-NN) architecture to perform effective intrusion detection. For this analysis, we utilize a comprehensive dataset consisting of data collected from diverse IIoT devices. The main contributions of this work include:
\begin{itemize}
    \item Performing data processing of the IIoT-based network intrusion detection dataset to enhance performance.
    \item Proposing an FTL-based approach to detect data breaches in IIoT environments, and introducing a novel neural network approach to perform the FTL.
    \item Analyzing and evaluating the performance of the proposed technique with traditional ML approaches. 
\end{itemize}

The rest of the paper is organized as follows: 
Section \ref{two} introduces the related work for this research. Our proposed methodology is shown in Section \ref{three}. Experimentation, results, and analysis are provided in Section \ref{four}. Finally, conclusions are drawn in Section \ref{five}. 

\section{Related Work} \label{two}
The development of network intrusion detection approaches has been a critical component in protecting today's networking infrastructures. With the emergence of IoT, these infrastructures need security contingencies now more than ever. Contemporary IoT NIDS surrounded the usage of signature-based mechanisms \cite{ioulianou2018signature} \cite{li2019designing} \cite{sheikh2018lightweight}. In \cite{ioulianou2018signature}, the authors proposed a signature-based NIDS that involved the usage of both centralized and distributed intrusion detection models. The work in \cite{li2019designing} developed CBSigIDS, a generic framework of collaborative blockchain signature-based intrusion detection systems, which can incrementally build and update a trusted signature database in a collaborative IoT environment. Researchers in \cite{sheikh2018lightweight} demonstrated the use of an optimized pattern recognition algorithm to detect such attacks within IoT. The limitation of signature-based methods is that they are static in operational capability and can be circumvented if attackers introduce a level of variability in their malware. 

With the rise of ML, ML-based NIDS has also become a trending technology to protect IoT systems \cite{tama2017attack} \cite{roy2018deep} \cite{chaabouni2019network}  \cite{alsaedi2020ton_iot} \cite{hindy2021machine}. The authors in \cite{tama2017attack} used a deep neural network to model various NIDS datasets to assess their performance at detecting attacks. This work did not take into account the impact of class imbalance on classifier performance. The work in \cite{roy2018deep} proposed a bi-directional long short-term memory neural network on a NIDS dataset to decipher between normal and attack conditions. The limitation of this work lay in the small size of the dataset that was used to create the model. Additionally, the trade-off between the decision parameters was not considered. Researchers in \cite{chaabouni2019network} applied ML to various open-sourced NIDS datasets. This work did not consider any fog or edge computing devices, which is not representative of real-life practices. In \cite{alsaedi2020ton_iot}, the researchers observed multiple ML models on an IoT dataset. Though this work formed the basis of contemporary research for NIDS for IoT environments, they did not perform any model hyperparameter optimization that impacts performance. In \cite{hindy2021machine}, authors evaluated six ML models for detecting MQTT-based IoT attacks. The knock on this work is that it only observed performance after training only on an MQTT protocol-based dataset. 

Additionally, to preserve privacy efforts in IoT, the usage of FL on IoT has seen some attention from researchers \cite{ferrag2022edge} \cite{attota2021ensemble} \cite{friha2022felids}. In \cite{ferrag2022edge}, the authors performed centralized FL on an IIoT dataset. The work done in \cite{attota2021ensemble} proposed an FL-based NIDS that trained on multiple views of IoT network data in a decentralized format to detect, classify, and defend against attacks. Lastly, the researchers in \cite{friha2022felids} proposed an FL-based intrusion detection system for securing agricultural-IoT infrastructures using local learning. The limitations of these works are that they have not used the advantages of the transfer learning, FL setup, which is the primary aim of this paper
.
\section{Methodology} \label{three}
\begin{figure*}[t]
\centerline{\includegraphics[width=16cm]{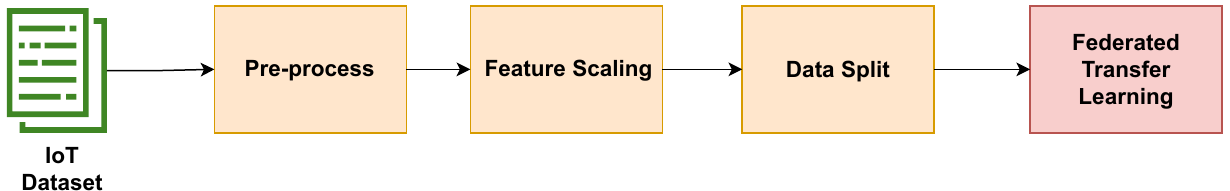}}
\caption{Proposed technique for the FTL Setup}
\label{Fig:approach}
\end{figure*}

The proposed approach for the FTL algorithms consists of a client-server architecture, where the models generated from the client data are utilized to update the server model in the setup. The steps of this proposed methodology are illustrated in Figure \ref{Fig:approach}.

\subsection{IoT Dataset}
The proposed approach has wide IIoT applications and is agnostic to dataset features. For this research, any open-sourced IoT dataset can be utilized for analysis. Let the dataset be denoted using the variable $Z$, consisting of $E$ total samples. $z_i$ represents a single data sample and $z_i \in Z, i = \{0,1...E\}$. After the aggregation of the dataset, the first step is to perform data pre-processing where redundant and/or incomplete information from the datasets can be eliminated to not hamper the performance of any trained ML model. Additionally, we also perform encoding in this step to ensure that no categorical features are being fed into the ML model. Once pre-processed, feature selection is performed to select the features that optimally affect the performance of the trained ML model. Following this, we split the pre-processed and feature-selected dataset into train and test samples. Finally, we parse the trained models into our FTL module.

\subsection{Pre-process}
In this stage, we evaluate the contents of the acquired datasets. Any features that seem to have redundancy or do not actively help in the classification task are removed. Examples of such features include data features where all values are the same in magnitude regardless of the labels associated with that data sample. Additionally, we also evaluate the dataset features and eliminate columns that consistently have ``infinity" or ``NaN" type values. These inputs will not be accepted by an ML classifier and can hamper system performance if used. Additionally, we also perform feature encoding in this stage, where categorical features are encoded to have numerical values to ensure that these inputs are accepted by ML classifiers. After data pre-processing, the original dataset $Z$ is now denoted by $Z'$. 

\subsection{Feature Scaling}
Once the dataset $Z'$ has been pre-processed, it moves into the feature scaling stage. The aim here is to identify dataset features that can be essential to the classification task and prioritize them, along with scaling them to fit within an operational range. To begin, we employ a correlation matrix. This is an algebraic and statistical tool that can be utilized to observe the relationship between two variables in a dataset. A correlation score is given for each pair of features which helps us to determine features that are not important. If it is positive, the features are directly correlated. If it is negative, the features are inversely correlated. The higher the magnitude of the score, the higher the correlation \cite{steiger1980tests}. We parse the dataset $Z'$ through the correlation matrix to unearth which data features are essential to the classification task. After analysis of the resultant matrix, the features with the lowest correlation scores get dropped. The threshold for dropping features is dependent on the variance of the correlation scores for the data features. 
These features get dropped as they result in higher cardinality for the dataset, while not providing any valuable insights into the data. Finally, we scale the resultant features to ensure that all feature magnitudes fit within a prescribed range. This makes sure that any trained model is not biased towards features that have higher magnitudes. This tactic is essential in datasets that have imbalanced magnitudes. Post feature scaling step, the resultant dataset is denoted by $X$.

\begin{figure*}
\centerline{\includegraphics[width=9cm]{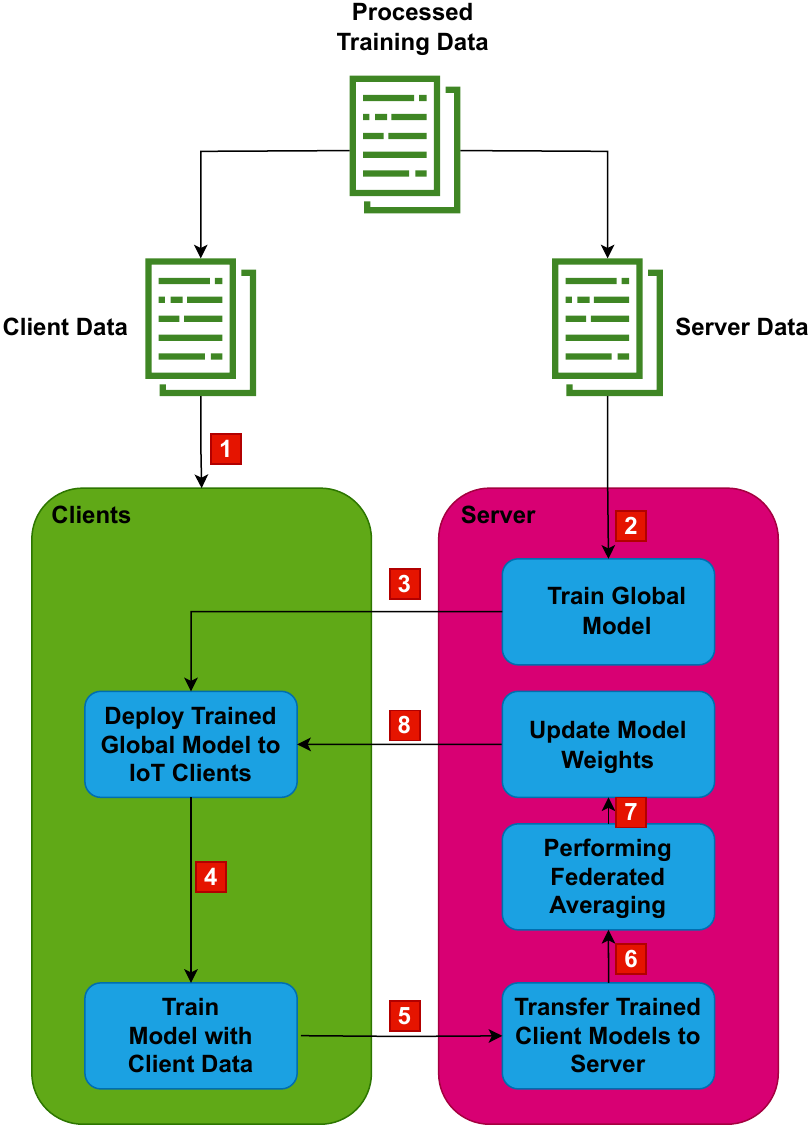}}
\caption{Overview and Steps in the FTL Process}
\label{Fig:ovr}
\end{figure*}

\subsection{Data Split}
After data collection, pre-processing, and feature selection, the full data suite $X$ is partitioned into the training $X_{train}$ and test $X_{test}$ sets. We assume the total number of training samples is $U$, while the total number of testing samples is $V$. It is also ensured that $U>V$.

\subsection{Federated Transfer Learning}
Once the data set has been split appropriately, the FTL approach can commence. In this situation, we are utilizing the combination of a deep neural network (DNN) and a convolutional neural network (CNN) as the base for our FTL in the IIoT setup. We shall refer to this machine learning architecture as a combinational neural network (Combo-NN). To conduct our FTL, we split the existing training data $X_{train}$ into two parts: the client data, denoted by $X_{c}$, and the server data, denoted by $X_{s}$. The proportion for splitting the data depends on multiple IoT characteristics like the size of the network and the number of devices that are in the IoT setup. Next, the server trains a Combo-NN with $X_{s}$. Once trained, the server models get deployed to all the IIoT clients. Within each client, the trained Combo-NN model gets retrained with the local client data. Once all the models are trained, clients send the models back to the server. The server performs federated averaging on these models and deploys the resultant model back to all the clients, and the procedure continues periodically till convergence. An overview of the FTL procedure is illustrated in Figure \ref{Fig:ovr}. The following sections highlight the important steps in the proposed approach. 

\begin{figure*}[t]
\centerline{\includegraphics[width=\textwidth]{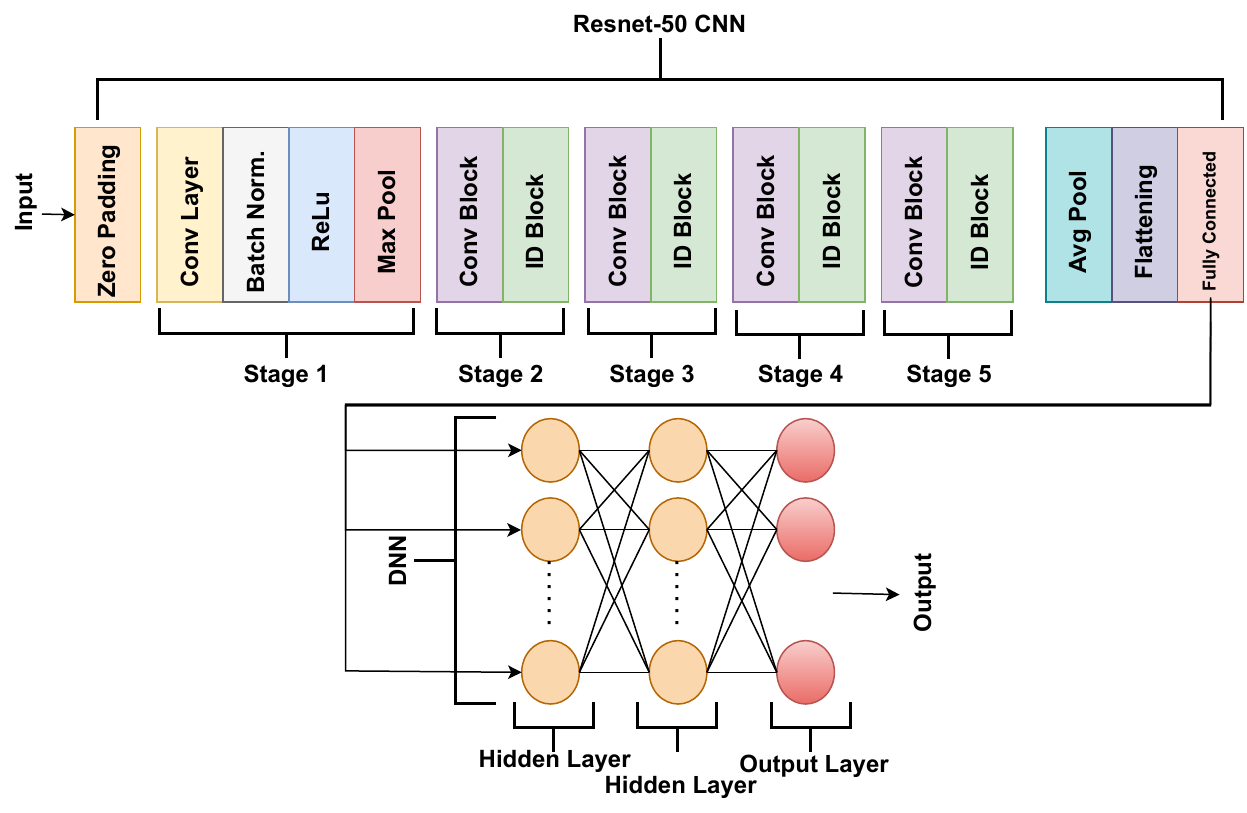}}
\caption{Proposed Combo-NN Architecture}
\label{Fig:combo}
\end{figure*}

\subsubsection{Client Data}
The client data, $X_{c}$, is then further split between all the clients in the IIoT FTL setup. Here we denote the number of IoT devices to be $N$, which each gets a subset of the client data. The local client data for each IoT client is denoted by $X_{c_i}$, where i represents an IoT client in the infrastructure. 

\subsubsection{Server Data and Model Training}
As mentioned above, the training data is also split into the server data, denoted by $X_{s}$. The server data is essential as this forms the basis for training our proposed Combo-NN model. The model trained on $X_{s}$ will be initially deployed to all of the client devices in the IoT infrastructure to begin the FTL procedure. 

The proposed Combo-NN architecture consists of a CNN and a DNN. In our methodology, we are using Resnet-50 as our CNN \cite{wen2020transfer}. We are utilizing Resnet-50 as it is an effective tool for generalizing nonlinear data in literature \cite{wen2020transfer}. However, other CNN architectures can also be utilized, depending on the proposed application of the FTL. CNNs can be beneficial for performing network intrusion detection as they tend to train fast due to their underlying architecture. Also, CNNs have the capability of training multi-layer networks with gradient descent that is competent to learn complex, high-dimensional, and nonlinear data \cite{aminanto2016deep}. Additionally, we are augmenting multiple hidden layers of a DNN to Resnet-50 for the proposed FTL approach. A DNN is chosen as it provides benefits to performing network intrusion detection as the layers of depth in the network are effective at modeling nonlinear interactions between the various observable features in the network \cite{covington2016deep}. The proposed Combo-NN architecture is illustrated in Figure \ref{Fig:combo}. 

Let our proposed untrained Combo-NN in the server be denoted by $M_s$. The network, then, gets trained using the server data $X_{s}$ by:
\begin{equation}
T_s = train(M_s, X_s)
 \end{equation}
where $T_s$ denotes the trained Combo-NN in the server device. After training, $T_s$ is duplicated $N$ times, each copy denoted by $T_{c_i}$, where $i$ refers to each of the $N$ IoT clients in the infrastructure. 

\subsection{Model Transfer to Clients}
After training the server model, $T_s$, in the previous stage, $T_s$ is duplicated $N$ times. Each duplicated copy is denoted by $T_{c_i}$, where $i$ refers to each of the $N$ IoT clients in the infrastructure. In this stage, all the trained duplicated copies are sent to all of the IoT client devices. Each client device deploys $T_{c_i}$ in their environment. 

\subsection{Model Training with Local Data}
After each model trained model $T_{c_i}$ is deployed in each IoT client $i$, in this stage, the model gets retrained using the local IoT data $X_{c_i}$. The purpose of this is to generalize the model with the local data for the IoT to provide better network intrusion detection performance for the client. The network gets trained using the client data by:
\begin{equation}
    M_{c_i} = train(T_{c_i}, X_{c_i})
\end{equation}

where $M_{c_i}$ denotes the trained Combo-NN model that was trained with the local client data of the IoT device. 

\subsection{Model Transfers to Server}
Once all the models in all $N$ clients are trained with the local data, these models are transferred to the IoT server. The purpose of this step is to accumulate all the trained models into a centralized location to perform the calibration of the global Combo-NN model located on the server. 

\subsection{Federated Averaging}
Once all of the trained models $M_c$ are transferred to the server, it can commence the federated averaging step to aggregate all the client models into one. FL is an iterative procedure and is traditionally conducted in multiple rounds. Here, $t$ denotes the number of rounds for the FL. In this stage, we perform the Federated Stochastic Gradient Descent (FedSGD) approach \cite{singh}. This is conducted through the following equation:
\begin{equation}
F_i(w_{t_s}) = \frac{1}{n_i}\sum_{j \in \rho_i}{} f_i(w_{t_s})
\end{equation}
where $w_{t_s}$ are the weights of the current server model $T_s$, $n_i$ represents the number of data points on the IoT client $i$, $\rho_i$ represents the set of data points on the IoT client $i$, $f_i$ denotes the loss that is achieved for the data sample ($x_i$, $y_i$) with the current model weights $w_{t_s}$. Next, the FedSGD step is finished using:
\begin{equation}
    g_{i} = Gradient (F_{i}(w_{t_{s_t}}))
\end{equation}
where $g_i$ is the computed gradients for every single IoT client model.

\subsection{Server Model Weight Updates}
The final step for our FTL approach is to update the weights of our server model based on the results achieved from the federated averaging step. In this stage, the central server aggregates the achieved gradients from the previous stage
and applies the update using:
\begin{equation}
    \forall i, w_{t_{s_{t+1}^i}} \gets w_{t_{s_{t}}} - \eta g_i
\end{equation}
Here, $\eta$ represents the learning rate. In this step, each client $i$ takes a step of gradient descent $g_i$ on the current server model using its local data with the learning rate $\eta$. The last step in this stage is:
\begin{equation}
    w_{t_{s_{t+1}}} \gets \sum_{i=1}^{N}\frac{n_i}{n} w_{t_{s_{t+1}^i}}
\end{equation}
In this step, the server takes the weighted average of the resulting models and updates the current server model in the system. 

\section{Experimentation and Results} \label{four}
\subsection{Experiment Setup}
For the experimentation, we utilized Jupyter Notebook as our development environment, along with Python being the development language. Some of the essential libraries used were pandas, scikit-learn, matplotlib, seaborn, keras, tensorflow, and psutil. The dataset that is used in our experiments is Edge-IIoTset Cyber Security Dataset of IoT \& IIoT \cite{ferrag2022edge}. For our experiment, we assume $N=2$. For evaluation, we are using Accuracy (A) and Macro-Average scores of Precision (MAP), Recall (MAR), and F-Measure (MAF). Additionally, we are evaluating the FTL approach against Logistic Regression (LR), Gaussian Naïve Bayes (GNB), Random Forest (RF), and Stochastic Gradient Descent (SGD) classifiers. These algorithms have been chosen as they are customarily used to study IoT network intrusion detection in the literature \cite{shitharth2022innovative} \cite{resende2018survey} \cite{abbasi2021anomaly}.

\subsection{Performance Evaluation and Analysis}
\begin{figure}[t]
\centerline{\includegraphics[width=\columnwidth]{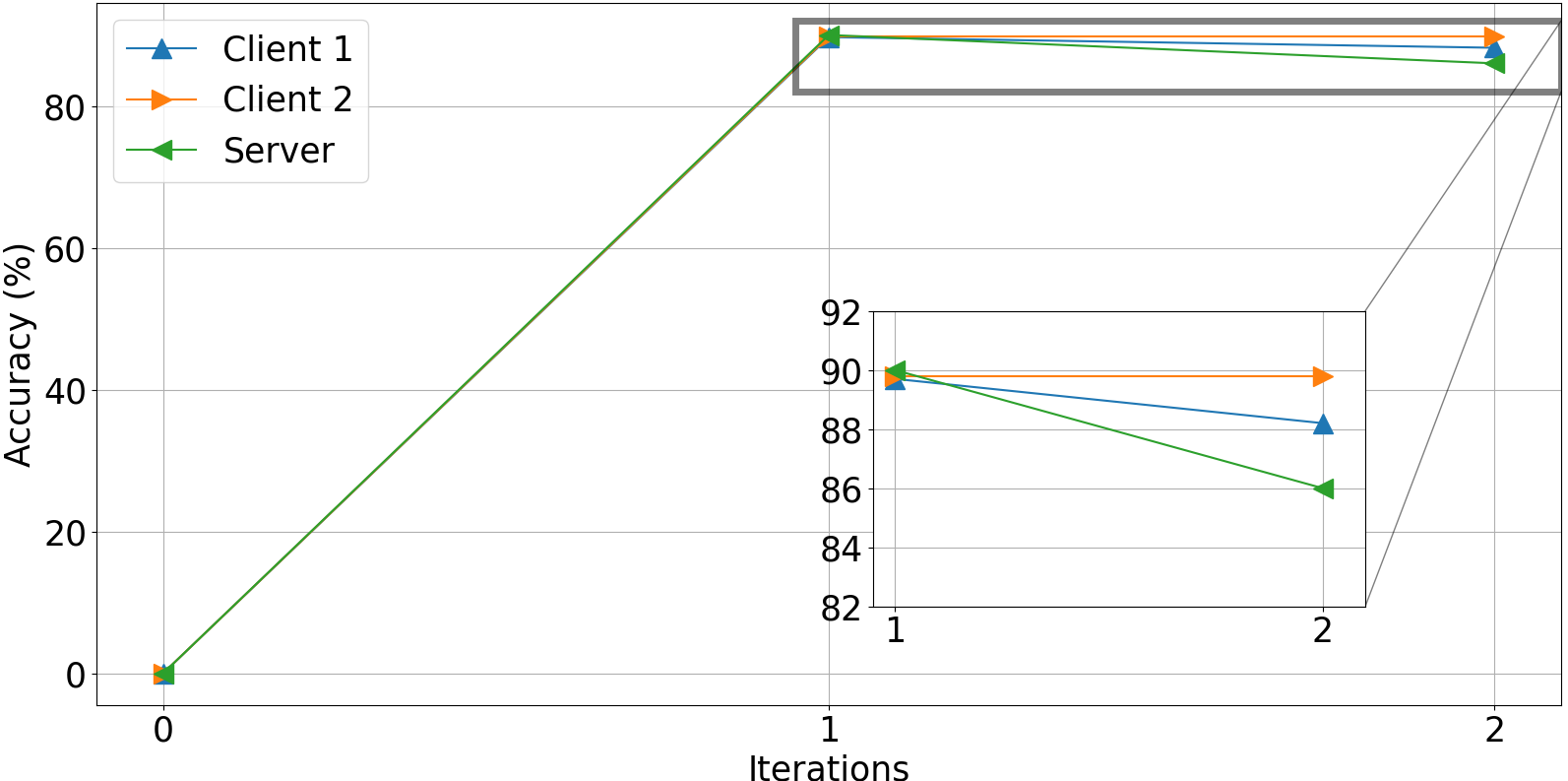}}
\caption{Accuracy Progression of Client and Server Performances through Iterations}
\label{Fig:client}
\end{figure}

First, we observe the performance achieved by the IIoT client models and the server model when they experience multiple iterations of the FTL approach. The results can be illustrated in \ref{Fig:client}. It should be noted that the server model's performance is dependent on that of the client models in the FTL setup. Here, we note that after the first iteration, Client 2 achieves a high accuracy of 89.8\% while Client 1 achieves a high accuracy of 89.7\%. Hence, after the first iteration, the server achieves high performance with an accuracy of 90\%. After the second iteration, we note that Client 2 maintains its performance with an 89.8\% accuracy. Client 1 slightly decreases in performance but still maintains a high accuracy score of 88.2\%. The server after the second iteration diminishes negligibly but achieves a high accuracy score of 86\%. In Client 1 and the server, we notice that slight decrease in performance between iterations, which could be attributed to multiple reasons like inter-client class imbalance and overall class imbalance in the dataset \cite{seol2023performance} \cite{thiyam2023efficient}.

\begin{figure}[t]
\centerline{\includegraphics[width=\columnwidth]{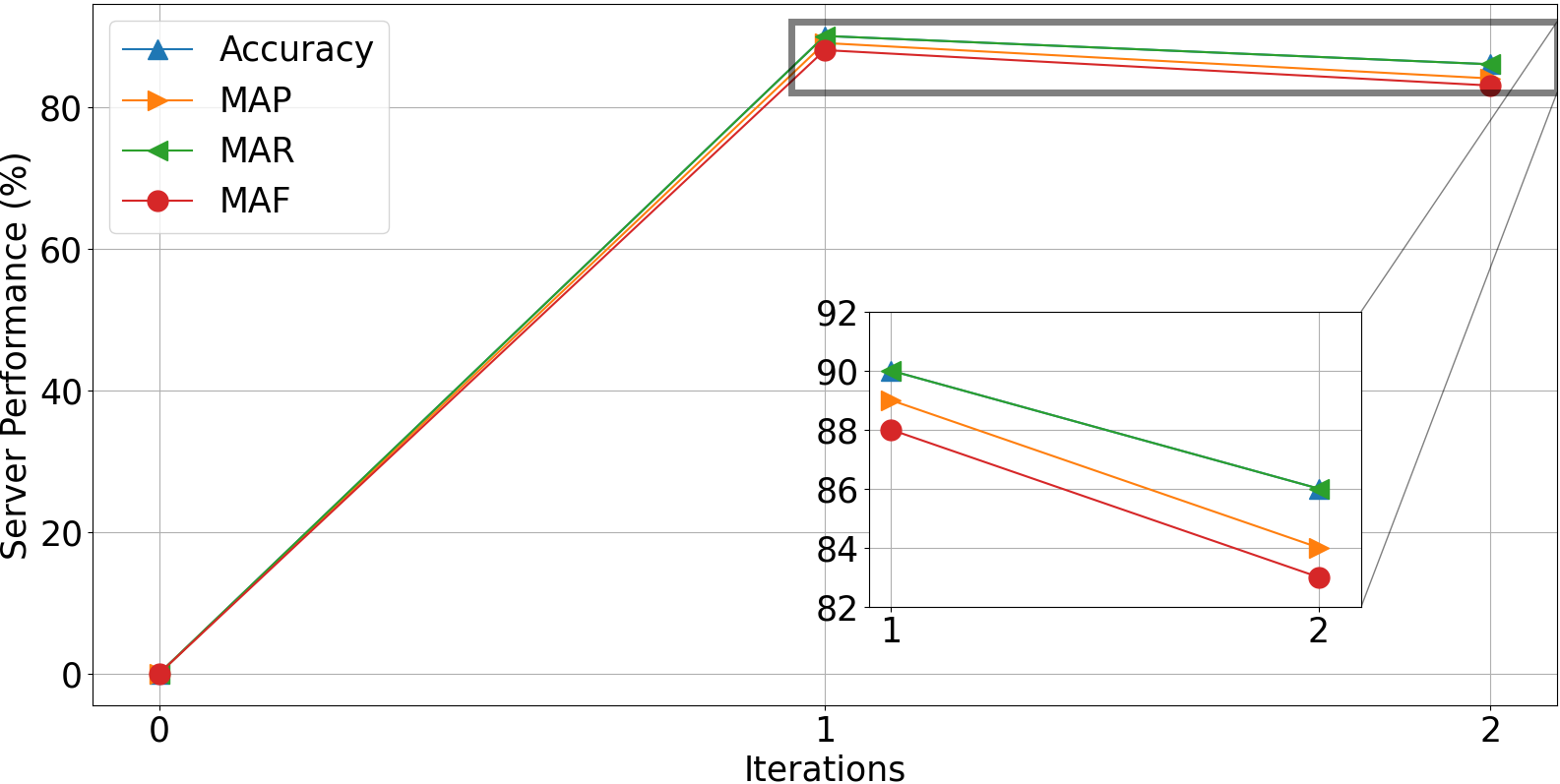}}
\caption{Progression of Server Performance through Iterations}
\label{Fig:server}
\end{figure}

Next, we evaluate the performance that the server IoT device experiences when undergoing the proposed FTL approach. The results from this experiment are illustrated in Figure \ref{Fig:server}. Here, we observe that when the server model has experienced the first iteration of the FTL approach, it achieves high performance with A, MAP, MAR, and MAF scores of 90\%, 89\%, 90\%, and 88\%, respectively. After the second iteration of the FTL approach, the FTL approach achieved comparable scores with A, MAP, MAR, and MAF scores of 86\%, 84\%, 86\%, and 83\%, respectively. We observe a slight decrease from the previous iteration, as previously noted in Figure \ref{Fig:client}. This can be attributed to multiple reasons like inter-client class imbalance and overall class imbalance in the dataset \cite{seol2023performance}.   

\begin{figure}[t]
\centerline{\includegraphics[width=\columnwidth]{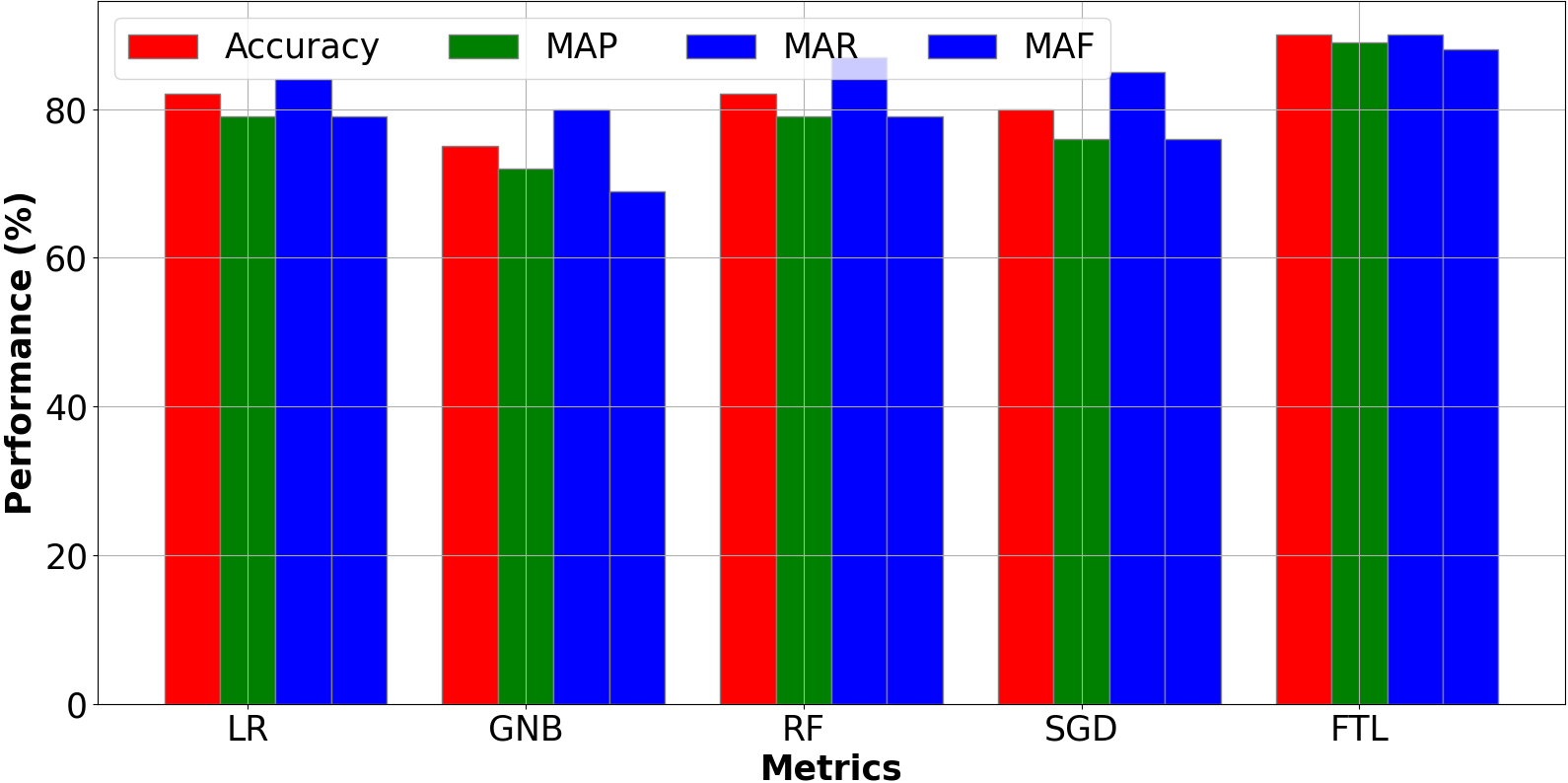}}
\caption{Performance Evaluation of the FTL against Traditional ML Algorithms}
\label{Fig:perf}
\end{figure}

Finally, we analyze the performance of the proposed FTL approach in performing multi-class classification. For this, we evaluate the effectiveness of the proposed approach against traditional ML algorithms that are customarily used for IoT network intrusion detection. The results are illustrated in Figure \ref{Fig:perf}. From the observed results, we notice that the GNB classifier achieves the least performance in being able to detect network intrusions with an \textit{A} score of 75\%, an MAP, MAR, and MAF scores of 72\%, 80\%, and 69\%, respectively. We also note that contemporary ML algorithms achieve better performance than the GNB classifier, but these achieved performances are not the optimal solutions. The performance achieved by the proposed Combo-NN-based FTL approach provides the best overall performance compared to the other contemporary ML algorithms. It achieves a 90\% accuracy score, and MAP, MAR, and MAF scores of 89\%, 90\%, and 88\% scores, respectively. This is because the FTL technique achieves optimal performance iteratively, making it more suitable for performing network intrusion detection over a wide set of network data. Additionally, the usage of a neural network-based architecture ensures that any deeply correlated network metrics that are not easily visible are identified by the neural network. This showcases that the proposed FTL techniques can effectively model the intrusion types in the dataset and better differentiate between the various labels. Therefore, using the proposed FTL approach can be a more effective mechanism to detect network intrusions in an IIoT setup versus other contemporary ML algorithms. 

\section{Conclusion} \label{five}
In this paper, we propose a new type of intrusion detection mechanism for Industrial IoT devices using the concept of Federated Transfer Learning. As part of this proposed work, we have introduced a novel neural network-based architecture called Combo-NN which conducts network intrusion detection utilizing a combination of a CNN and DNN. To evaluate the efficacy of our proposed approach, we analyze its performance on an open-source IIoT network intrusion detection dataset. Results showcase high performance for the FTL setup between iterations on both the IIoT clients and the server. Additionally, we observe that the proposed FTL setup can achieve better overall performance than contemporary ML algorithms that have previously been used to perform network intrusion detection in the literature. To the best of our knowledge, this is the first time FTL is being used to perform network intrusion detection for IIoT. Future work will focus on analyzing the performance of the proposed technique to other network intrusion detection datasets, along with developing potential security features to protect the approach from adversarial attacks. 

\bibliographystyle{IEEEtran}
\bibliography{biblio}

\begin{thebibliography}{10}
\providecommand{\url}[1]{#1}
\csname url@samestyle\endcsname
\providecommand{\newblock}{\relax}
\providecommand{\bibinfo}[2]{#2}
\providecommand{\BIBentrySTDinterwordspacing}{\spaceskip=0pt\relax}
\providecommand{\BIBentryALTinterwordstretchfactor}{4}
\providecommand{\BIBentryALTinterwordspacing}{\spaceskip=\fontdimen2\font plus
\BIBentryALTinterwordstretchfactor\fontdimen3\font minus
  \fontdimen4\font\relax}
\providecommand{\BIBforeignlanguage}[2]{{%
\expandafter\ifx\csname l@#1\endcsname\relax
\typeout{** WARNING: IEEEtran.bst: No hyphenation pattern has been}%
\typeout{** loaded for the language `#1'. Using the pattern for}%
\typeout{** the default language instead.}%
\else
\language=\csname l@#1\endcsname
\fi
#2}}
\providecommand{\BIBdecl}{\relax}
\BIBdecl

\bibitem{cisco2020cisco}
U.~Cisco, ``Cisco annual internet report (2018--2023) white paper,''
  \emph{Cisco: San Jose, CA, USA}, vol.~10, no.~1, pp. 1--35, 2020.

\bibitem{cyrus2021iot}
C.~Cyrus, ``Iot cyberattacks escalate in 2021, according to kaspersky,''
  \emph{IoT World Today, https://www. iotworldtoday.
  com/2021/09/17/iot-cyberattacks-escalate-in-2021-according-to-kaspersky/,
  last accessed December}, 2021.

\bibitem{al2021review}
R.~Al-amri, R.~K. Murugesan, M.~Man, A.~F. Abdulateef, M.~A. Al-Sharafi, and
  A.~A. Alkahtani, ``A review of machine learning and deep learning techniques
  for anomaly detection in iot data,'' \emph{Applied Sciences}, vol.~11,
  no.~12, p. 5320, 2021.

\bibitem{sharma2021optimal}
N.~V. Sharma and N.~S. Yadav, ``An optimal intrusion detection system using
  recursive feature elimination and ensemble of classifiers,''
  \emph{Microprocessors and Microsystems}, vol.~85, p. 104293, 2021.

\bibitem{zhang2021survey}
C.~Zhang, Y.~Xie, H.~Bai, B.~Yu, W.~Li, and Y.~Gao, ``A survey on federated
  learning,'' \emph{Knowledge-Based Systems}, vol. 216, p. 106775, 2021.

\bibitem{zhuang2020comprehensive}
F.~Zhuang, Z.~Qi, K.~Duan, D.~Xi, Y.~Zhu, H.~Zhu, H.~Xiong, and Q.~He, ``A
  comprehensive survey on transfer learning,'' \emph{Proceedings of the IEEE},
  vol. 109, no.~1, pp. 43--76, 2020.

\bibitem{ioulianou2018signature}
P.~Ioulianou, V.~Vasilakis, I.~Moscholios, and M.~Logothetis, ``A
  signature-based intrusion detection system for the internet of things,''
  \emph{Information and Communication Technology Form}, 2018.

\bibitem{li2019designing}
W.~Li, S.~Tug, W.~Meng, and Y.~Wang, ``Designing collaborative blockchained
  signature-based intrusion detection in iot environments,'' \emph{Future
  Generation Computer Systems}, vol.~96, pp. 481--489, 2019.

\bibitem{sheikh2018lightweight}
N.~U. Sheikh, H.~Rahman, S.~Vikram, and H.~AlQahtani, ``A lightweight
  signature-based ids for iot environment,'' \emph{arXiv preprint
  arXiv:1811.04582}, 2018.

\bibitem{tama2017attack}
B.~A. Tama and K.-H. Rhee, ``Attack classification analysis of iot network via
  deep learning approach,'' \emph{Res. Briefs Inf. Commun. Technol.
  Evol.(ReBICTE)}, vol.~3, pp. 1--9, 2017.

\bibitem{roy2018deep}
B.~Roy and H.~Cheung, ``A deep learning approach for intrusion detection in
  internet of things using bi-directional long short-term memory recurrent
  neural network,'' in \emph{2018 28th international telecommunication networks
  and applications conference (ITNAC)}.\hskip 1em plus 0.5em minus 0.4em\relax
  IEEE, 2018, pp. 1--6.

\bibitem{chaabouni2019network}
N.~Chaabouni, M.~Mosbah, A.~Zemmari, C.~Sauvignac, and P.~Faruki, ``Network
  intrusion detection for iot security based on learning techniques,''
  \emph{IEEE Communications Surveys \& Tutorials}, vol.~21, no.~3, pp.
  2671--2701, 2019.

\bibitem{alsaedi2020ton_iot}
A.~Alsaedi, N.~Moustafa, Z.~Tari, A.~Mahmood, and A.~Anwar, ``Ton\_iot
  telemetry dataset: A new generation dataset of iot and iiot for data-driven
  intrusion detection systems,'' \emph{Ieee Access}, vol.~8, pp.
  165\,130--165\,150, 2020.

\bibitem{hindy2021machine}
H.~Hindy, E.~Bayne, M.~Bures, R.~Atkinson, C.~Tachtatzis, and X.~Bellekens,
  ``Machine learning based iot intrusion detection system: An mqtt case study
  (mqtt-iot-ids2020 dataset),'' in \emph{Selected Papers from the 12th
  International Networking Conference: INC 2020}.\hskip 1em plus 0.5em minus
  0.4em\relax Springer, 2021, pp. 73--84.

\bibitem{ferrag2022edge}
M.~A. Ferrag, O.~Friha, D.~Hamouda, L.~Maglaras, and H.~Janicke,
  ``Edge-iiotset: A new comprehensive realistic cyber security dataset of iot
  and iiot applications for centralized and federated learning,'' \emph{IEEE
  Access}, vol.~10, pp. 40\,281--40\,306, 2022.

\bibitem{attota2021ensemble}
D.~C. Attota, V.~Mothukuri, R.~M. Parizi, and S.~Pouriyeh, ``An ensemble
  multi-view federated learning intrusion detection for iot,'' \emph{IEEE
  Access}, vol.~9, pp. 117\,734--117\,745, 2021.

\bibitem{friha2022felids}
O.~Friha, M.~A. Ferrag, L.~Shu, L.~Maglaras, K.-K.~R. Choo, and M.~Nafaa,
  ``Felids: Federated learning-based intrusion detection system for
  agricultural internet of things,'' \emph{Journal of Parallel and Distributed
  Computing}, vol. 165, pp. 17--31, 2022.

\bibitem{steiger1980tests}
J.~H. Steiger, ``Tests for comparing elements of a correlation matrix.''
  \emph{Psychological bulletin}, vol.~87, no.~2, p. 245, 1980.

\bibitem{wen2020transfer}
L.~Wen, X.~Li, and L.~Gao, ``A transfer convolutional neural network for fault
  diagnosis based on resnet-50,'' \emph{Neural Computing and Applications},
  vol.~32, pp. 6111--6124, 2020.

\bibitem{aminanto2016deep}
E.~Aminanto and K.~Kim, ``Deep learning in intrusion detection system: An
  overview,'' in \emph{2016 International Research Conference on Engineering
  and Technology (2016 IRCET)}.\hskip 1em plus 0.5em minus 0.4em\relax Higher
  Education Forum, 2016.

\bibitem{covington2016deep}
P.~Covington, J.~Adams, and E.~Sargin, ``Deep neural networks for youtube
  recommendations,'' in \emph{Proceedings of the 10th ACM conference on
  recommender systems}, 2016, pp. 191--198.

\bibitem{singh}
\BIBentryALTinterwordspacing
S.~Singh, ``\BIBforeignlanguage{en-us}{{PPML} {Series} \#2 - {Federated}
  {Optimization} {Algorithms} - {FedSGD} and {FedAvg}},'' Dec. 2021. [Online].
  Available:
  \url{https://shreyansh26.github.io/post/2021-12-18_federated_optimization_fedavg/}
\BIBentrySTDinterwordspacing

\bibitem{shitharth2022innovative}
S.~Shitharth, P.~R. Kshirsagar, P.~K. Balachandran, K.~H. Alyoubi, and A.~O.
  Khadidos, ``An innovative perceptual pigeon galvanized optimization (ppgo)
  based likelihood na{\"\i}ve bayes (lnb) classification approach for network
  intrusion detection system,'' \emph{IEEE Access}, vol.~10, pp.
  46\,424--46\,441, 2022.

\bibitem{resende2018survey}
P.~A.~A. Resende and A.~C. Drummond, ``A survey of random forest based methods
  for intrusion detection systems,'' \emph{ACM Computing Surveys (CSUR)},
  vol.~51, no.~3, pp. 1--36, 2018.

\bibitem{abbasi2021anomaly}
F.~Abbasi, M.~Naderan, and S.~E. Alavi, ``Anomaly detection in internet of
  things using feature selection and classification based on logistic
  regression and artificial neural network on n-baiot dataset,'' in \emph{2021
  5th International Conference on Internet of Things and Applications
  (IoT)}.\hskip 1em plus 0.5em minus 0.4em\relax IEEE, 2021, pp. 1--7.

\bibitem{seol2023performance}
M.~Seol and T.~Kim, ``Performance enhancement in federated learning by reducing
  class imbalance of non-iid data,'' \emph{Sensors}, vol.~23, no.~3, p. 1152,
  2023.

\bibitem{thiyam2023efficient}
B.~Thiyam and S.~Dey, ``Efficient feature evaluation approach for a
  class-imbalanced dataset using machine learning,'' \emph{Procedia Computer
  Science}, vol. 218, pp. 2520--2532, 2023.

\end{thebibliography}

\end{document}